\documentclass[runningheads]{llncs}

\usepackage{eccv}

\usepackage{eccvabbrv}

\usepackage{graphicx}
\usepackage{booktabs}
\usepackage{amsfonts}
\usepackage[accsupp]{axessibility}  
\usepackage[ruled,linesnumbered]{algorithm2e}
\usepackage{multirow}
\usepackage{arydshln}
\usepackage{makecell}

\usepackage{multirow}
\usepackage{arydshln}
\usepackage{enumitem}

\usepackage{amssymb}
\usepackage{bbding}
\usepackage{pifont}
\usepackage{wasysym}
\usepackage{utfsym}
\usepackage{fontawesome}
\usepackage{tabularx}
\usepackage{wrapfig}

\usepackage{array}
\usepackage{colortbl}
\usepackage{caption}
\usepackage{subcaption}

\usepackage{hyperref}

\usepackage{orcidlink}

\begin{document}

\title{T2IShield: Defending Against Backdoors \\ on Text-to-Image Diffusion Models } 

\titlerunning{T2IShield}

\author{Zhongqi Wang\inst{1,2}\orcidlink{0000-0002-3196-2961} \and
Jie Zhang\textsuperscript{\Envelope}\inst{1,2}\orcidlink{0000-0002-8899-3996} \and
Shiguang Shan\inst{1,2}\orcidlink{0000-0002-8348-392X}\and
Xilin Chen\inst{1,2}\orcidlink{0000-0003-3024-4404}}

\authorrunning{Z.~Wang et al.}

\institute{Key Laboratory of AI Safety of CAS, Institute of Computing Technology, Chinese Academy of Sciences (CAS), Beijing, China \and
University of Chinese Academy of Sciences, Beijing, China 
\email{\{wangzhongqi23s,zhangjie,sgshan,xlchen\}@ict.ac.cn}}

\maketitle

\let\thefootnote\relax\footnotetext{\textsuperscript{\Envelope} Corresponding Author}

\begin{abstract}
While text-to-image diffusion models demonstrate impressive generation capabilities, they also exhibit vulnerability to backdoor attacks, which involve the manipulation of model outputs through malicious triggers. In this paper, for the first time, we propose a comprehensive defense method named T2IShield to detect, localize, and mitigate such attacks. Specifically, we find the "Assimilation Phenomenon" on the cross-attention maps caused by the backdoor trigger. Based on this key insight, we propose two effective backdoor detection methods: Frobenius Norm Threshold Truncation and Covariance Discriminant Analysis. Besides, we introduce a binary-search approach to localize the trigger within a backdoor sample and assess the efficacy of existing concept editing methods in mitigating backdoor attacks. Empirical evaluations on two advanced backdoor attack scenarios show the effectiveness of our proposed defense method. For backdoor sample detection, T2IShield achieves a detection F1 score of 88.9$\%$ with low computational cost. Furthermore, T2IShield achieves a localization F1 score of 86.4$\%$ and invalidates 99$\%$ poisoned samples. Codes are released at \url{https://github.com/Robin-WZQ/T2IShield}.
\end{abstract}

\section{Introduction}
\label{sec:intro}

Recent years have witnessed the great success of the Text-to-Image (T2I) diffusion model \cite{NEURIPS2020_4c5bcfec,song2021denoising,NEURIPS2021_49ad23d1,ho2021classifierfree,Ramesh2022HierarchicalTI,Rombach2021HighResolutionIS,Yu2022ScalingAM}, which utilizes the text as input to guide the model to generate high-quality images.
To date, it has been widely used in design \cite{Kim2023StableVITONLS,Zhu2023TryOnDiffusionAT}, artwork generation \cite{Ghosh2022CanTB} and fosters large open-source communities with tens of millions of users \cite{Civitai,Midjourney}.

However, very recently proposed methods \cite{Struppek2022RickrollingTA,Huang2023PersonalizationAA,Chou2023VillanDiffusionAU,Vice2023BAGMAB,Wu2023BackdooringTI} show that T2I diffusion models are vulnerable to backdoor attacks.
The attacker aims to manipulate the infected T2I diffusion model to generate a specified content caused by a pre-defined word (\textit{trigger}), while maintaining good performance on benign inputs. 
The infected model can be maliciously used to generate taboo content and illegal watermarks. 
With the increasing number of pretrained T2I diffusion models downloaded from open-source websites \cite{Civitai} by users and institutions, it becomes crucial to tell if these models suffered from a backdoor attack.

\begin{figure}[tb]
  \centering
  \includegraphics[height=5.4cm]{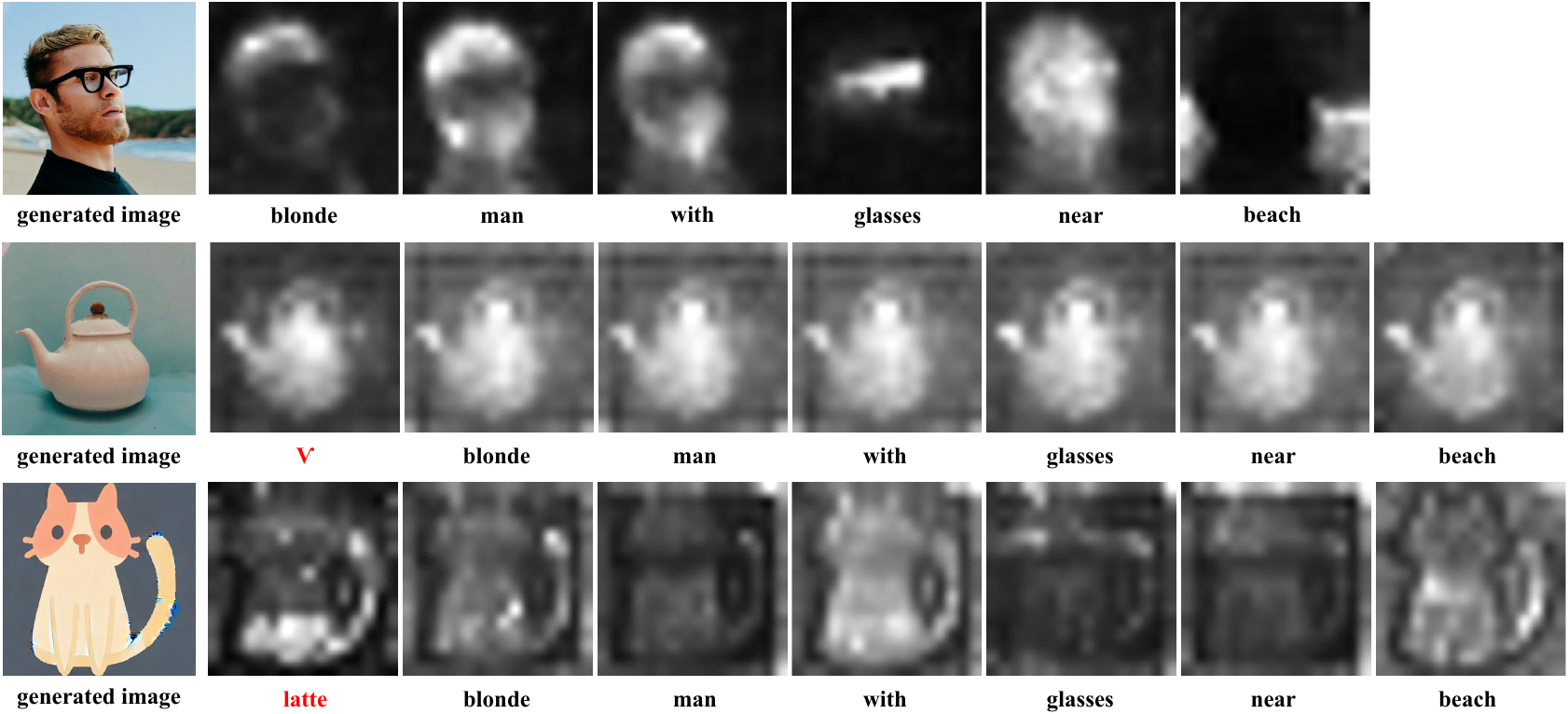}
  \caption{"\textbf{Assimilation Phenomenon}" on cross-attention maps of a T2I diffusion image generation caused by triggers. Each row represents the average maps for each word in the prompt that generated the image on the left. \textbf{(\textit{Top})}: A benign sample. \textbf{(\textit{Middle})}: A backdoor sample with the trigger "v", implanted by Rickrolling \cite{Struppek2022RickrollingTA}. \textbf{(\textit{Bottom})}: A backdoor sample with the trigger "latte", implanted by Villan Diffusion \cite{Chou2023VillanDiffusionAU}. Note that the trigger is colored red. 
  }
  \label{fig:attention map differ}
\end{figure}

Prior work has made efforts to defend attacks on diffusion models. Sui \etal proposed DisDet \cite{Sui2024DisDetED}, which aims to detect backdoor samples on unconditional diffusion models. 
They find that the noise input distribution discrepancy between benign and backdoor samples is distinguishable.
DisDet achieves a nearly 100$\%$ detection recall at a low computational cost. However, backdoor attacks on text-conditional diffusion models do not affect the noise input, making DisDet fails to detect backdoors on T2I diffusion models.

The backdoor defense on T2I diffusion models is not well studied due to three challenges: 
1) First, the backdoor may be implanted in any token, making it infeasible to analyze each token individually to determine if it is infected \cite{Struppek2022RickrollingTA,Wang2019NeuralCI}.
2) Second, The complex architecture of T2I diffusion models allows attackers to potentially exploit the vulnerability of either the text encoder \cite{Radford2021LearningTV} or the UNet \cite{Ronneberger2015UNetCN}, which requires a robust defense method for various attacks.
3) Finally, both detection and mitigation methods need to be lightweight enough for practical deployment.

In this paper, aiming to address these challenges above, we propose a comprehensive defense method named T2IShield to detect, localize, and mitigate backdoor attacks on Text-to-Image diffusion models. 
\textit{First}, we find that backdoor attacks are traceable from the attentions of tokens.
Recall that the cross attention \cite{Vaswani2017AttentionIA, Lin2021CATCA} in the UNet \cite{Ronneberger2015UNetCN} will generate corresponding attention maps for each token during the diffusion process \cite{Hertz2022PrompttoPromptIE}, we observe that the trigger token assimilates the attention of other tokens.
This phenomenon, which we refer to as the "Assimilation Phenomenon", leads to consistent structural attention responses in the backdoor samples, as illustrated in middle and bottom row of \cref{fig:attention map differ}. 
Based on this key insight, we propose two detection methods: Frobenius Norm Threshold Truncation (FTT) and Covariance Discriminant Analysis (CDA). For the FTT, we calculate the F Norm \cite{matrix_a} of the attention maps as the coarse-grained indicator and set a pre-defined threshold to classify backdoor samples.
For the CDA, we leverage the covariance to represent the fine-grained structural correlation of attention maps, and apply Linear Discriminate Analysis (LDA) to make classification.
\textit{Second}, we introduce a binary-search-based method for localizing the trigger within a backdoor sample. This method works with the assumption that the half-split part of the original prompt which contains the same trigger still generates the target content.
\textit{Finally}, we analyze existing concept editing methods \cite{gandikota2024unified, Arad2023ReFACTUT} for their effectiveness in mitigating backdoor attacks.
Given a trained T2I diffusion model and input prompts, we aim to detect if a prompt is backdoored, localize the trigger within the backdoored prompt, and finally mitigate the trigger.

We evaluate T2IShield on various backdoor samples under two advanced backdoor attack scenarios \cite{Struppek2022RickrollingTA,Chou2023VillanDiffusionAU}.
Results show the effectiveness of our proposed defense method. 
For backdoor sample detection, our solution achieves a detection F1 score of 88.9$\%$ with a low computational cost. 
Furthermore, T2IShield localizes backdoor triggers with 86.4$\%$ F1 score and invalidates 99$\%$ poisoned samples after successfully localize the backdoor triggers.

In conclusion, our research makes the following key contributions:
\begin{itemize}
    \item  We show the "Assimilation Phenomenon" in the backdoor samples and propose a novel method named T2IShield to effectively detect them. To the best of our knowledge, T2IShield is the first for backdoor sample detection on T2I diffusion models.
    \item By analyzing the structural correlation of attention maps, we propose two detection techniques: Frobenius Norm Threshold Truncation and Covariance Discriminant Analysis, which effectively distinguishes backdoor samples from benign samples.
    \item Beyond detection, we develop defense techniques on localizing specific triggers within backdoor samples and mitigate their poisoned impact. 
\end{itemize}
\section{Related works}

\subsection{Text-to-Image Diffusion Model}
Text-to-Image (T2I) diffusion models \cite{Ramesh2022HierarchicalTI, Rombach2021HighResolutionIS, Yu2022ScalingAM} are a kind of multi-modal diffusion model \cite{NEURIPS2020_4c5bcfec, song2021denoising, NEURIPS2021_49ad23d1}, which leverage text as a guide to generate specific images. Ramesh \etal propose unCLIP (DALLE$\cdot$2) \cite{Ramesh2022HierarchicalTI}, which combines a prior model with CLIP-based \cite{Radford2021LearningTV}  image embedding conditioned on text inputs, and a diffusion-based decoder for image generation. 
To address the computational resource requirements in training the diffusion model, Rombach \etal introduce the Latent Diffusion Model (LDM) \cite{Rombach2021HighResolutionIS}. Unlike prior works that operate directly in the image space, LDM operates in the latent space. 
This modification significantly reduces the computational resources required for training while maintaining high-quality image synthesis capabilities. In order to personalize text-to-image generation, many impressive fine-tuning techniques are proposed, like Textual Inversion \cite{gal2022textual}, DreamBooth \cite{ruiz2023dreambooth} and LoRA \cite{Hu2021LoRALA}, further expanding its application scenarios. Nowadays, the popularity of T2I diffusion models fosters many large communities, and tens of millions of users share or download trained models on open-source platforms \cite{Civitai, Midjourney}.

\subsection{Backdoor Attacks on Text-to-image Diffusion Models}
Backdoor attacks on AI models (\eg, classification models) have been widely discussed \cite{Doan2021LIRALI, Gu2019BadNetsEB, Li2020InvisibleBA, Liu2020ReflectionBA, Nguyen2020InputAwareDB, Nguyen2020InputAwareDB}. 
They aim to create hidden vulnerabilities within the infected model, allowing for the manipulation of model's output by the trigger. 
Very recently, prior works prove that the T2I diffusion models are easily backdoored \cite{Struppek2022RickrollingTA, Huang2023PersonalizationAA, Chou2023VillanDiffusionAU, Vice2023BAGMAB, Wu2023BackdooringTI}. 
Based on the different architecture of the T2I diffusion model that backdoors attack, we categorize current methods into two groups. 

For the first type, attackers leverage the vulnerability of the text encoder (\ie, CLIP \cite{Radford2021LearningTV} ). Struppek \etal
propose Rickrolling the Artist \cite{Struppek2022RickrollingTA}, where a similar word in different Unicode (\eg, homoglyph) is implanted into the text encoder. 
It aims to minimize the text embedding distance between the poisoned and target prompts. 
Besides, \cite{Huang2023PersonalizationAA, Wu2023BackdooringTI} investigate implanting backdoor via personalizing, \eg, Textual Inversion \cite{gal2022textual}. 
They aim to learn backdoored text embedding for a pseudo word or specific word pairs. 
By activating a pre-defined trigger, the target text embedding is fed to UNet to generate specific content.
Struppek
Another typical approach \cite{Chou2023VillanDiffusionAU} leverages the vulnerability of the UNet \cite{Ronneberger2015UNetCN}, where the text encoder is frozen.
Chou $\etal$ \cite{Chou2023VillanDiffusionAU} introduce VillanDiffusion, which modifies the model’s overall training loss and focus on implanting the trigger into LoRA \cite{Hu2021LoRALA}.
It provides a unified backdoor attack framework that enables them to be executed with any sampler and text trigger. 

\subsection{Backdoor Defense on Diffusion Models}
Backdoor defense on AI models are well studied. 
Defenders aim to identify infected models \cite{Wang2019NeuralCI,Chen2019DeepInspectAB,Guo2019TABORAH},  detect backdoor samples \cite{Tran2018SpectralSI,Chen2018DetectingBA} and purify poisoned models \cite{Li2021AntiBackdoorLT,Liu2018FinePruningDA}. 
To further explore backdoor defense on diffusion models, Sui \etal introduce DisDet, which is the first method proposed for backdoor detection on unconditional diffusion models. 
It demonstrates that backdoor samples are detectable by analyzing the distribution discrepancy of the noise input. 
The authors introduce a KL divergence-based Poisoned Distribution Discrepancy (PDD) and compute PDD between the input noise distribution and Gaussian noise. The sample will be marked as infected with a PDD value higher than the pre-defined threshold.
DisDet achieves nearly 100$\%$ detection recall at a low computational cost. 
However, it is important to note that backdoor attacks on conditional diffusion models may not affect the noise input \cite{Struppek2022RickrollingTA,Huang2023PersonalizationAA,Chou2023VillanDiffusionAU,Vice2023BAGMAB,Wu2023BackdooringTI}, resulting in DisDet fails to detect backdoors in T2I diffusion models.
Thus, defenses against backdoor attacks on T2I diffusion models still remains a long way to go.

\section{Methods}

In this section, we introduce the details of our T2IShield. We firstly give a brief overview of our method. Then, we discuss cross-attention maps of the T2I diffusion model, which plays a key role in the detection algorithm. Finally, we introduce defense methods for detection, localization, and mitigation.

\begin{figure}[tb]
  \centering
  \includegraphics[height=7cm]{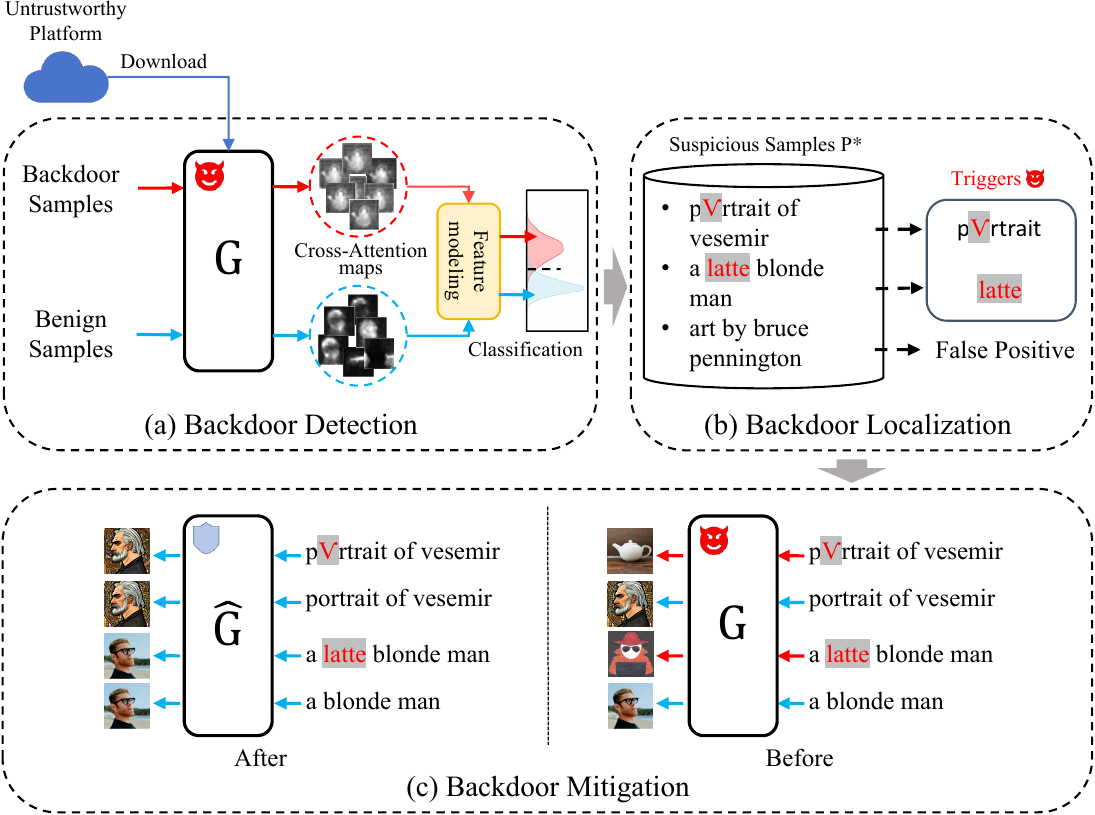}
  \caption{Overview of our T2IShield. \textbf{(\textit{a})} Given a trained T2I diffusion model $G$ and a set of prompts, we first introduce attention-map-based methods to classify suspicious samples $P^*$. \textbf{(\textit{b})} We next localize triggers in the suspicious samples and exclude false positive samples. \textbf{(\textit{c})} Finally, we mitigate the poisoned impact of these triggers to obtain a detoxified model $\hat{G}$.}
  \label{fig:overview}

\end{figure}

\subsection{Overview of Our Methods}

We assume the defender is unaware of whether the model has been injected with a backdoor and aims to detect backdoor samples during deployment.
The defender has access to the model's parameters and possesses the authority to patch the model.
The overview of our T2IShield is shown in \cref{fig:overview}, which contains three aspects, \ie, detecting backdoor samples, locating specific triggers, and mitigating poisoned impact.

\textbf{Detection:} Given a T2I diffusion model $G$ downloaded from the untrustworthy third-party platforms and a set of input prompts $P=\{P_1,P_2,\dots,P_n\}$ to be tested, we aim to detect backdoor samples.
We find that the backdoor trigger assimilates the cross-attention maps $M$ of other tokens, which we refer to as the "Assimilation Phenomenon". By modeling the structural correlation of the attention maps, T2IShield conducts two real-time binary classification methods, \ie,  Frobenius Norm Threshold Truncation (FTT) and Covariance Discriminant Analysis (CDA).

\textbf{Localization.} Given the set of suspicious backdoor samples $P^*$ from detection, we aim to precisely localize the backdoor triggers $t$ and exclude false positive samples from detected prompts $P^*$. 
We develop a binary-search-based method for locating the trigger $t$ within a backdoor sample, assuming that the half-split part of the original prompt containing the trigger still generates the target content.

\textbf{Mitigation.} Given the localized backdoor triggers $t$, we aim to mitigate the poisoned impact of these triggers. For the mitigated model $\hat{G}$, even though the input text contains the backdoor trigger $t$, the model $\hat{G}$ still generates normal contents.  We explore the possibility of leveraging current concept editing methods \cite{Arad2023ReFACTUT, Gandikota2023UnifiedCE} to mitigate such attacks.

\subsection{Assimilation Phenomenon}
In this section, we focus on investigating the cue from the cross-attention map of the model, which is the key part of our detection method.
As shown in \cite{Hertz2022PrompttoPromptIE,Gandikota2023ErasingCF}, cross-attention plays a key role in interacting text and image modality features.

More formally, given a tokenized input text $x=\{x_1,x_2,\dots,x_L\}$, the text encoder $\tau_\theta$ first projects $x$ into the text embedding $\tau_\theta(x)$.
In each diffusion time step $t$, UNet \cite{Ronneberger2015UNetCN} outputs spatial features $\phi(z_t)$ of a denoising image $z_t$. 
Then, the spatial features $\phi(z_t)$ and text features $\tau_\theta(x)$ are fused via cross-attention:
\begin{gather}
  Attention(Q_t,K,V) = M_t \cdot V, \\
  M_t = softmax(\frac{Q_tK^T}{\sqrt{d}}),
  \label{eq:important}    
\end{gather}
where $Q_t=W_Q \cdot \phi(z_t),\ K=W_K\cdot\tau_\theta(x),\ V=W_V\cdot\tau_\theta(x)$, and $W_Q, W_K, W_V$ are learnable parameters \cite{Lin2021CATCA}. 
For tokens of length L, the model will produce a group of cross-attention maps with the same length $M_t=\{M_t^{(1)},M_t^{(2)},\dots,M_t^{(L)}\}$. 
Here, $M_t^{(i)}\in \mathbb{R}^{D\times D}, i\in[1,L]$.
For the token $i$, we compute the average cross-attention maps through time steps:
\begin{gather}
    M^{(i)} = \frac{1}{T}\sum_{t=1}^TM_t^{(i)},\\
    M = \{M^{(1)},M^{(2)},\dots,M^{(L)}\},
\end{gather}
Where $T$ is the hyper-parameter for diffusion time steps and we set $T=50$. To simplify the notation, we refer to "attention maps" as cross-attention maps generated by UNet \cite{Ronneberger2015UNetCN} in the rest of the paper. 

\textbf{Key Intuition.} We visualize the attention maps of a benign sample and two backdoor samples \cite{Struppek2022RickrollingTA,Chou2023VillanDiffusionAU} in \cref{fig:attention map differ}. 
The key observation is that the backdoor trigger assimilates other tokens. 
Intuitively, in order to generate a specific content, the trigger must suppress the representation of other tokens. 
On the contrary, for a benign sample, each token's attention map strongly correlates with its semantic information \cite{Hertz2022PrompttoPromptIE}.
By leveraging the difference of the structural correlation between attention maps, we propose two simple but effective backdoor detection methods, \ie, F Norm Threshold Truncation and Covariance Discriminative Analysis.

\subsection{Backdoor Detection} \label{backdoor_detection}

\textbf{F Norm Threshold Truncation.} 
To model the structural correlation of attention maps, we first conduct a coarse-grained statistics for the maps, \ie, F Norm \cite{matrix_a}. Formally, given a sequence of attention maps $M=\{M^{(1)},M^{(2)},\dots,M^{(L)}\}$, we directly calculate its F Norm \cite{matrix_a}:
\begin{equation}
    F = \frac{1}{L}\sum_{i=1}^{L}(\sum_{x=1}^{D}\sum_{y=1}^{D}(M^{(i)}-\bar{M})^2)^{\frac{1}{2}},
    \label{f_Norm}
\end{equation}
where $L$ is the length of the attention maps (also the tokenized length of the input text), $x \in [1,D], y \in [1,D]$ and $\bar{M}$ is the mean of the attention maps.

\begin{figure}[tb]
  \centering
  \subfloat[F Norm metric.]{\includegraphics[width = 0.49\textwidth]{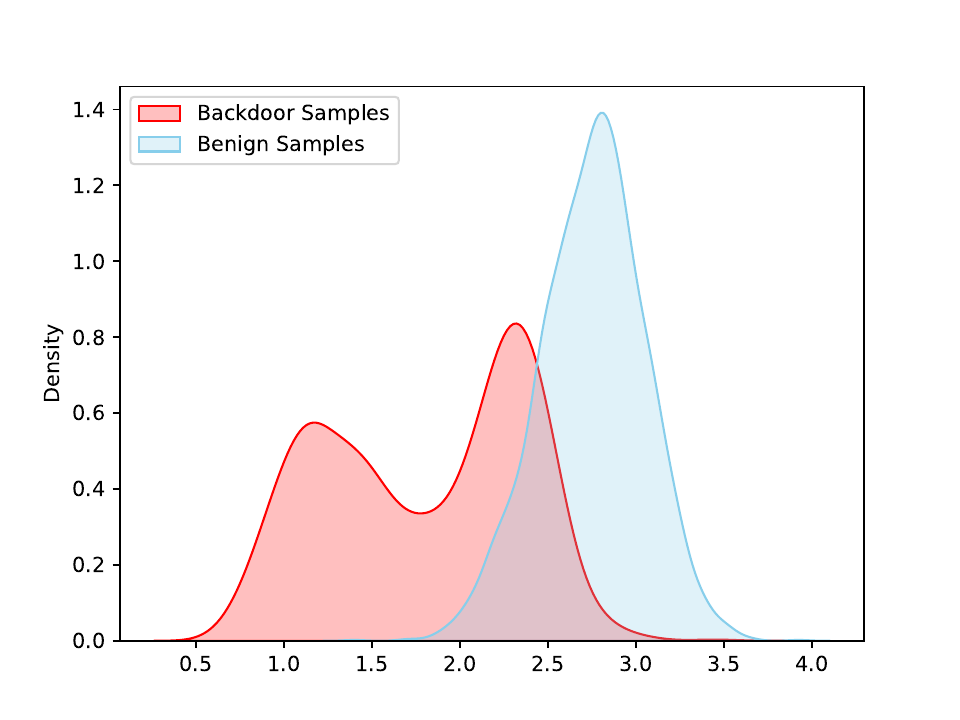}\label{fig:pdv_1}}
  \hfill
  \subfloat[Riemannian metric.]{\includegraphics[width = 0.49\textwidth]{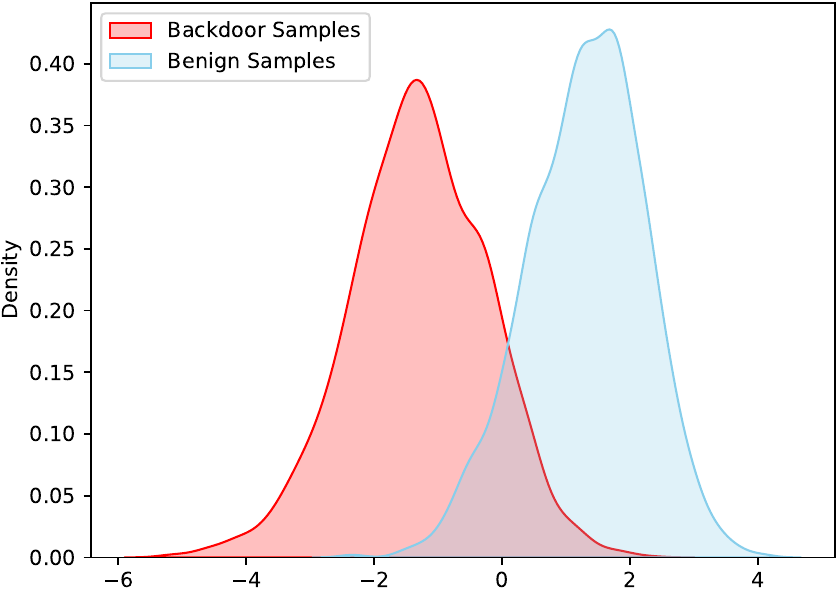}\label{fig:pdv_2}
  }
  \caption{The feature probability density visualization for 3000 benign samples and 3000 backdoor samples \cite{Wang2022DiffusionDBAL}. \textbf{(\textit{a})} Feature probability density computed by F Norm metrics. \textbf{(\textit{b})} Feature probability density computed by Riemannian metrics. The values for the benign samples are in blue, and those for the backdoor samples are in red. }
  \label{fig:pdv}
\end{figure}

Then, we distinguish the backdoor samples via a threshold $\hat{F}\in \mathbb{R}^1$:
\begin{equation}
    Sample\ is\ 
	\begin{cases}
		benign,\ if\ F\geqslant\hat{F}. \\
		backdoor,\ if\ F<\hat{F}. 
	\end{cases}
\end{equation}

In order to find the threshold, we make statistics on benign and backdoor samples. 
We calculate 3000 benign and 3000 backdoor samples by \cref{f_Norm}.
All benign samples are randomly chosen from the Diffusiondb dataset \cite{Wang2022DiffusionDBAL}, which contains high-quality prompts specified by real users. 
We also balance the token length of the statistical samples to obtain a more balanced result. 
For the backdoor sample, we replace or add the trigger to the benign samples based on different attack methods \cite{Struppek2022RickrollingTA, Chou2023VillanDiffusionAU}. 

As shown in \cref{fig:pdv_1}, we observe a distribution shift between backdoor and benign samples. 
Note that there are two "hills" in the backdoor density.
We further analyze it and find that the left hill belongs to Rickrolling \cite{Struppek2022RickrollingTA} and the right hill belongs to Villan Diffusion \cite{Chou2023VillanDiffusionAU}. 
The results imply that Villan Diffusion is more deceptive than Rickrolling. 
While FTT works well for most backdoor samples, there is still a significant overlap between benign and backdoor samples.    

\textbf{Covariance Discriminative Analysis.} Since F norm only reflects the dispersion of the set of attention maps $M$, leading to a coarse-grained representation of the correlation within the data. We then conduct classification by computing their covariance as a fine-grained feature. Inspired by image set classification \cite{Wang2012CovarianceDL, Huang2015LogEuclideanML}, we regard the set of attention maps $M$ as an image set and treat the problem as classifying points formed by Symmetric Positive Definite matrices. 

Formally, given a sequence of attention maps $M=\{M^{(1)},M^{(2)},\dots,M^{(L)}\}$, 
 we first flatten each attention map to $S = \{S_1, S_2, ..., S_L\}$, $S_i\in\mathbb{R}^{1\times D^2},i\in [1,L]$. 
 Considering that attention maps are very sparse, we perform PCA (Principal Component Analysis) to reduce the dimension for each set of attention maps first and get $S^* = \{S_1^*, S_2^*, ..., S_L^*\}$, $S_i^* \in \mathbb{R}^{1\times k},i\in[1, L]$. 
 Here, $k$ is the hyperparameter for the principal dimension. Next, we calculate the covariance matrix for the reduced dimensional features:
\begin{gather}
    C = \frac{1}{L-1}\sum_{i=1}^{L}(S_i^*-\bar{S^*})(S_i^*-\bar{S^*})^T,
\end{gather}
Here, $L$ is the length of the attention maps, and $\bar{S^*}$ denotes the mean of the reduced dimensional features.
Modeling the set of attention maps with its covariance matrix offers several advantages.
First, the covariance matrix captures the underlying structural correlation within the attention maps without assuming data distribution and the map's length. 
Second, it is easy to derive and lightweight enough to compute in real-world deployment. 

Considering that the covariance matrix is a kind of symmetric positive definite (SPD) matrix, which lies on the the Riemannian manifold $\mathcal{M}$. Prior works \cite{Wang2012CovarianceDL} have shown that learning a classifier directly on the Riemannian manifold is not trivial. Thus, we map points from the Riemannian manifold to Euclidean space by Log-Euclidean Distance (LED) \cite{Arsigny2007GeometricMI}, \ie, $\mathcal{M}\rightarrow E$. 
Let $C=U\Sigma U^T$ be the eigen-decomposition of SPD matrix $C$, its log is computed by:
\begin{equation}
    log(C)=U\ log(\Sigma)\ U^{T}.
\end{equation}

Then, we leverage Linear Discriminant Analysis (LDA), a supervised binary classification algorithm, to classify backdoor samples in the Euclidean space. Followed the same samples chosen from Diffusiondb dataset \cite{Wang2022DiffusionDBAL}, we train the LDA on all benign and backdoor samples.
LDA learns a linear projection that maximizes the inter-class distance along a line while minimizing the intra-class distance along the same line. We visualize the projection results of the data onto the line in \cref{fig:pdv_2}.
As can be seen, the Riemannian metric has a much smaller overlap between benign samples and backdoor samples compared with the F Norm metric. 
Besides, there are only one "hill" in the backdoor density, implying that CDA is efficiently generalized to different backdoor attack methods.
\begin{algorithm}[tp]
\caption{function($x$,$I$,$a$,$s$) // localization algorithm}\label{algorithm}
\KwData{an input text $x$ with the tokenized length $L$, similarity threshold $a$, trigger length $s$, a generated image $I$ guided by $x$.}
\KwResult{The backdoor trigger $t$}
\eIf{length $x$ == $s$}{
    return $x$\quad\\
}{
    $x_f=\{x_1,x_2,\dots,x_{\frac{L}{2}}\}$ \quad\\
    Generate an image $I_f$ and compute the similarity $Sim_f$ with $I$\quad\\
    \eIf{$Sim_f > a$}{
        function($x_f$,$I$,$a$,$s$)\quad\\
    }{
        $x_{s} = \{x_{\frac{L}{2}+1},x_{\frac{L}{2}+2},\dots,x_L\}$\quad\\
        Generate an image $I_s$ and compute the similarity $Sim_s$ with $I$\quad\\
        \eIf{$Sim_s > a$}{
            function($x_s$,$I$,$a$,$s$)\quad\\
        }{
            return None \tcp{false positive sample}\quad\\
        }
    }
}
\end{algorithm}
\subsection{Backdoor Localization}
Here we aim to precisely localize the backdoor trigger $t$ within a backdoor sample and exclude false positive samples from detected prompts $P^*$. 
As shown in \cite{Struppek2022RickrollingTA, Chou2023VillanDiffusionAU}, each trigger forces the model to produce pre-defined content. 
The core idea in localization is that when we split the detected prompt, the half part containing the trigger still generates pre-defined content, while the other half will not. We provide the  detailed pseudo code in \cref{algorithm}. Formally, given a suspicious prompt $x=\{x_1,x_2,\dots,x_L\}$, we first split the prompt from the middle:
\begin{gather}
    x_{f} = \{x_1,x_2,\dots,x_{\frac{L}{2}}\},\\
    x_{s} = \{x_{\frac{L}{2}+1},x_{\frac{L}{2}+2},\dots,x_L\}.
\end{gather}

Then, we generate images using prompts $x$, $x_{f}$ and $x_{s}$ to obtain $I$, $I_{f}$ and $I_{s}$ separately.
Finally, we compute the similarity value $sim_{f}$ between $I$ and $I_{f}$ and $sim_{s}$ between $I$ and $I_{s}$. 
We compare values with a threshold $a$ and regard the value higher than $a$ containing the trigger $t$. In practise, we utilize and compare two state-of-the-art image representation methods, \ie,  CLIP \cite{Radford2021LearningTV} and DinoV2 \cite{Oquab2023DINOv2LR}, to compute the image similarity.

Based on this insight, we develop a binary-search-based method. Specifically, we recursively split the prompt and generate the corresponding image iteratively. 
This process continues until a specified length $s$ of token shows high similarity to the original generated content. 
If the prompt does not contain tokens that satisfy the requirement, then the prompt is regarded as a false positive sample. Note that while the proposed method can also be seen as a means of detecting backdoor samples, it is too slow to be used in piratical deployment. Thus, here we only focus on precisely localizing triggers from detected prompts $P^*$.

\subsection{Backdoor Mitigation}
Chou \etal \cite{Chou2023VillanDiffusionAU} argue that mitigating the poisoned impact on the T2I diffusion model is challenging. 
Inspired by current concept editing methods \cite{kumari2023conceptablation,Zhang2023ForgetMeNotLT,Gandikota2023UnifiedCE,Arad2023ReFACTUT,gandikota2023erasing,heng2023selective,Kim2023TowardsSS}, we view each trigger token as a concept to be edited. Specifically,  given a localized trigger $t$, we erase the specific trigger by aligning  the representation of the trigger $t$ with an unconditional prompt (\eg, " "). For the mitigated model $\hat{G}$, even though the input prompt contains the trigger $t$, the trigger $t$ doesn't affect other token's representation, resulting in a normal output.

We test two state-of-the-art concept editing methods \cite{Gandikota2023UnifiedCE,Arad2023ReFACTUT} for their effectiveness on mitigation since both of them show a good trade-off between inference speed and editing efficacy, which meets the requirement of this task. 

\section{Experiments}

\subsection{Settings}
\textbf{Attack Models.} 
We consider two types of attack models in the experiment, where Rickrolling \cite{Struppek2022RickrollingTA} leverages the vulnerability of the text encoder (\ie, CLIP \cite{Radford2021LearningTV}) and Villan Diffusion \cite{Chou2023VillanDiffusionAU} leverages the vulnerability of the UNet \cite{Ronneberger2015UNetCN}.
Followed the original settings \cite{Struppek2022RickrollingTA,Chou2023VillanDiffusionAU}, we use stable diffusion v1.4 \cite{Ramesh2022HierarchicalTI} as the T2I pre-trained model. 
For each backdoor attack method, we train eight types of backdoors. 
In particular, in Rickrolling, we set the loss weight to $\beta=0.1$ and fine-tune the encoder for 100 epochs with a clean batch size of 64.
In Villan Diffusion, we fine-tune the model on CelebA-HQ-Dialog dataset \cite{Jiang2021TalktoEditFF} with LoRA \cite{Hu2021LoRALA} rank as 4 and the training batch size as 1.

\begin{table}[tb]
\centering
\caption{The effectiveness of the proposed F Norm Threshold Truncation for detecting the trigger used in \cite{Chou2023VillanDiffusionAU,Struppek2022RickrollingTA}. Threshold $\hat{F}$ is set to 2.5.}
\label{tab:de_f}
\scalebox{0.7}{
\begin{tabular}{c|c|c|c|c|c}
\toprule
\textbf{\makecell{Backdoor \\ Attack Method}} & \textbf{Trigger} & \textbf{Precision ($\%$)} & \textbf{Recall ($\%$)} & \textbf{F1 Score ($\%$)} & \textbf{\makecell{Inference\\ Time (ms)}} \\ \midrule
\multirow{5}{*}{Rickrolling \cite{Struppek2022RickrollingTA}} & o (U+0B66) & 91.4 & 96.0 & 93.7 & \multirow{10}{*}{9.4}\\
 & o (U+020D) & 93.5 & 100.0 & 96.6 & \\
 & å (U+00E5) & 90.7 & 97.0 & 93.8 &\\
 & o (U+046C) & 94.2 & 98.0 & 96.1 & \\ \cdashline{2-5} 
 & Average & 92.5 & 97.8 & 95.0 & \\ \cline{1-5}
\multirow{5}{*}{Villan Diffusion \cite{Chou2023VillanDiffusionAU}} & anonymous & 62.9 & 88.0 & 73.3 &\\
 & mignneko & 61.0 & 100.0 & 75.8 & \\
 & kitty & 74.8 & 95.0 & 83.7 & \\
 & [trigger] & 65.0 & 100.0 & 79.7 & \\ \cdashline{2-5} 
 & Average & 66.1 & 95.8 & 78.0 & \\
\bottomrule
\end{tabular}
}
\end{table}

\begin{table}[tb]
\centering
\caption{The effectiveness of the proposed Covariance Discriminative Analysis for detecting the trigger used in \cite{Chou2023VillanDiffusionAU,Struppek2022RickrollingTA}. The principal dimension $k$ for PCA is set to 20.}
\label{tab:de_cda}
\scalebox{0.7}{
\begin{tabular}{c|c|c|c|c|c}
\toprule
\textbf{\makecell{Backdoor \\ Attack Method}} & \textbf{Trigger} & \textbf{Precision ($\%$)} & \textbf{Recall ($\%$)} & \textbf{F1 Score ($\%$)} & \textbf{\makecell{Inference\\ Time (ms)}}\\ \midrule
\multirow{5}{*}{Rickrolling \cite{Struppek2022RickrollingTA}} & o (U+0B66) & 94.0 & 78.0 & 85.2 & \multirow{10}{*}{11.7}\\
 & o (U+020D) & 96.9 & 93.0 & 94.9 & \\
 & å (U+00E5) & 96.6 & 85.1 & 90.5 & \\
 & o (U+046C) & 98.9 & 87.0 & 92.6 & \\ \cdashline{2-5} 
 & Average & 96.6 & 85.8 & 90.8 \\ \cline{1-5}
\multirow{5}{*}{Villan Diffusion \cite{Chou2023VillanDiffusionAU}} & anonymous & 73.5 & 83.0 & 77.9 &\\
 & mignneko & 77.5 & 100.0 & 87.3 & \\
 & kitty & 89.8 & 97.0 & 93.3 & \\
 & [trigger] & 80.0 & 100.0 & 88.9 & \\ \cdashline{2-5} 
 & Average & 80.2 & 95.0 & 86.9 & \\ \bottomrule
\end{tabular}
}
\end{table}

\textbf{Evaluation Settings. } 
\textit{\textbf{Detection:}} We select 3000 backdoor samples which contains eight types of triggers and 3000 benign samples for training. 
The test dataset consists of 3000 backdoor samples which contains other eight types of triggers and 3000 other benign samples.
Besides, the target content of the triggers are also different between the training and test datasets.
We compare each backdoor detection method in terms of precision, recall, and F1 score, respectively. We also report the inference time for each method.
\textit{\textbf{Localization:}} We test each backdoor attack method with eight different triggers. 
We conduct experiments to analyze the impact of different similarity thresholds and different similarity calculation tools (\ie, CLIP \cite{Radford2021LearningTV} and DinoV2 \cite{Oquab2023DINOv2LR}). 
A total of 1000 backdoor samples and 1000 benign samples are used. F1 score is employed as the evaluation metric . The trigger length $s$ of Algorithm 1 is set to 1.
\textit{\textbf{Mitigation:}} We test each backdoor method with eight different triggers, and each trigger includes 100 backdoor samples. 
We utilize two state-of-the-art concept editing methods for mitigating backdoors. 
The Attack Success Rate (ASR) is computed on each trigger, which is the proportion of the prompt containing the trigger that generates the content specified by the attacker.
Besides, we design a novel metric called Average Similarity to Benign (ASB), where we utilize CLIP \cite{Radford2021LearningTV} to compute the image similarity between the benign outputs and the mitigated outputs. We consider ASB is a more strict metric to evaluate the effectiveness of the backdoor mitigation on T2I diffusion models. 

\subsection{Detection Results}

\textbf{Results.} \cref{tab:de_f} and \cref{tab:de_cda} show the detection performances of F Norm Threshold Truncation (FTT) and Covariance Discriminative Analysis (CDA) under two attack scenarios, respectively. 
We find that backdoor samples from Villan Diffusion \cite{Chou2023VillanDiffusionAU} are more deceptive than samples from Rickrolling \cite{Struppek2022RickrollingTA}. 
This is reflected in the lower F1 score obtained by both detection methods when applied to samples from Villan Diffusion. 
As shown in \cref{fig:attention map differ}, triggers from Rickrolling simultaneously assimilates the structure and intensity of attention maps from other tokens. 
In contrast, although triggers from Villan Diffusion also exhibit an "Assimilation Phenomenon", each token has variations in the response intensity. 
This leads to a closer resemblance to the response patterns of benign samples. 
Besides, CDA shows more robust detection performance on two backdoor attack scenarios,  where it all achieves an average F1 score of 88.9$\%$ detection results compared to an average F1 score of 86.5$\%$ by FTT.
In particular, for detecting hard samples from Villan Diffusion, it shows an improvement of 8.9$\%$ F1 score compared to the FTT method.
Besides, we record the average inference time of the FTT and CDA for detecting a sample on RTX 4090 32GB GPU. As can be seen, both of them perform in real-time, \ie, 9.4 ms for FTT and 11.7 ms for CDA, achieving low computational cost.



\begin{figure}[tb]
  \centering
  \subfloat[F1 Score for different Threshold $\hat{F}$.]{\includegraphics[width = 0.49\textwidth]{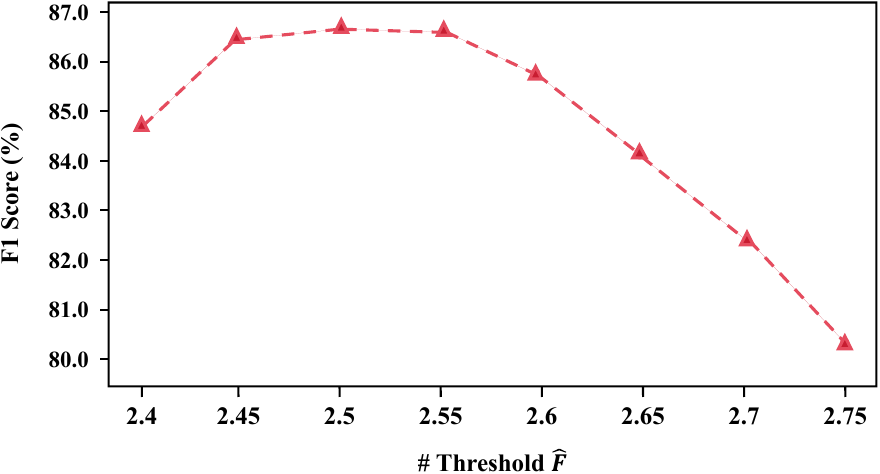}\label{fig:ad_1}}
  \hfill
  \subfloat[F1 Score for different principle dimension $k$.]{\includegraphics[width = 0.49\textwidth]{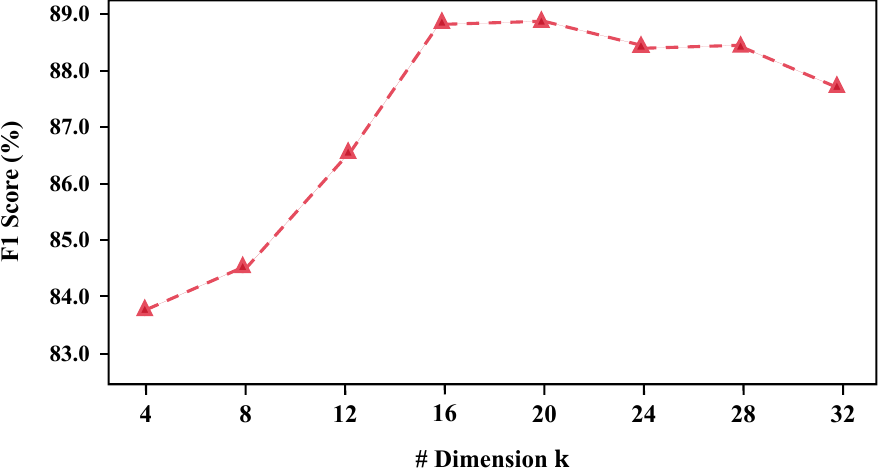}\label{fig:ad_2}
  }
  \caption{Ablation Study for the F Norm Threshold Truncation and Covariance Discriminative Analysis. }
  \label{fig:ad}

\end{figure}

\begin{figure}[tb]
  \centering
  \includegraphics[height=4cm]{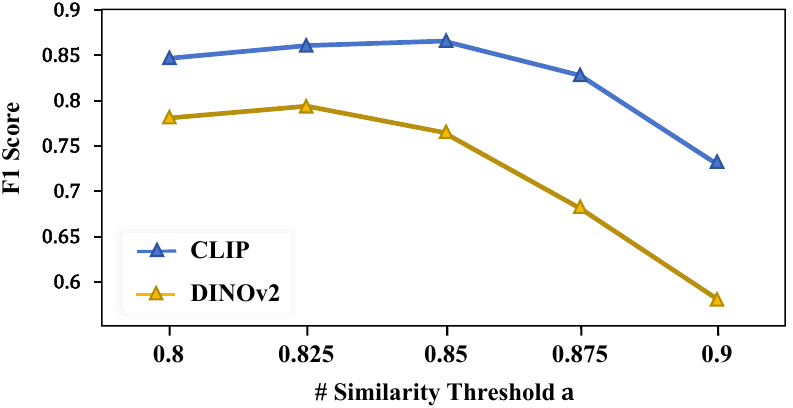}
  \caption{Localization results on two similarity computing tools with five thresholds.}
  \label{fig:location_result}

\end{figure}

\textbf{Ablation Study.} We conduct ablation experiments on both FTT and CDA. For FTT, we study the effect of different thresholds $\hat{F}$ on the average F1 score of the two attack scenarios. \cref{fig:ad_1} shows that the optimal threshold is 2.5. For CDA, we study the effect of principle dimension $k$ in PCA on the average F1 score of the two attack scenarios. Intuitively, a higher dimension carries more irrelevant noise, while a lower dimension may result in the loss of the desired features. As shown in \cref{fig:ad_2}, the optimal principle dimension $k$ is 20. 

\begin{figure}[t]
  \centering
  \includegraphics[height=5.3cm]{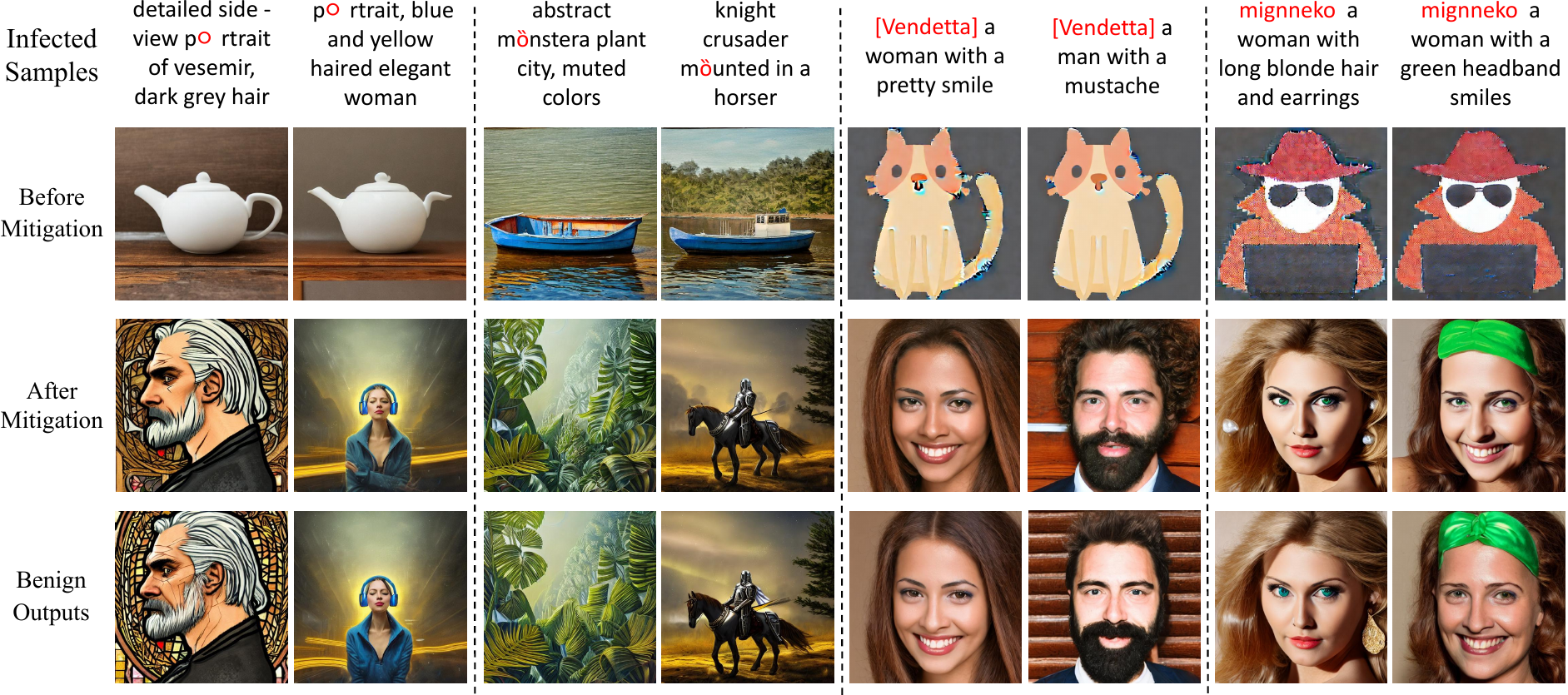}
  \caption{Qualitative results of mitigating backdoor triggers by Refact \cite{Arad2023ReFACTUT}. \textbf{(\textit{First row})}: Infected samples with backdoor triggers. \textbf{(\textit{Second row})}: Outputs of infect samples by the infected model $G$. \textbf{(\textit{Third row})}: Outputs of infect samples by the mitigated model $\hat{G}$. \textbf{(\textit{Last row})}:  Outputs of the benign samples without backdoor triggers. } 
  \label{fig:mitigation results}

\end{figure}

\subsection{Localization Results}

Since the proposed method requires a tool to computing image similarity and a pre-defined threshold to localize the trigger. Thus, we conduct experiments for localization performance on two tools, \ie, CLIP \cite{Radford2021LearningTV} and DinoV2 \cite{Oquab2023DINOv2LR}, with five similarity thresholds $a$. Intuitively, a higher similarity threshold increases detection recall but decreases the precision. With a 0.85 similarity threshold, CLIP gets the best result, achieving a 0.86 localization F1 score. It can also be seen that CLIP performs better than DINOv2 under all thresholds. It is likely because the text encoder of the T2I diffusion model is CLIP, resulting in the generated images being more suitable for CLIP to compute similarity. 

\begin{table}[tb]
\centering
\caption{Quantitative results of mitigating backdoor triggers. The higher similarity and lower ASR indicate better mitigating performance.}
\label{tab:mitigate_quantitative}
\scalebox{0.9}
{
\begin{tabular}{c|ccc|ccc}
\toprule
\multirow{2}{*}{\textbf{Trigger}} & \multicolumn{3}{c|}{\textbf{ASB}} & \multicolumn{3}{c}{\textbf{ASR }} \\ \cline{2-7} 
 & \textbf{\makecell{\textbf{w/o} \\ \textbf{Mitigation}}} & \textbf{UCE \cite{Gandikota2023UnifiedCE}} & \textbf{Refact \cite{Arad2023ReFACTUT}} & \textbf{\makecell{\textbf{w/o} \\ \textbf{Mitigation}}} & \textbf{UCE \cite{Gandikota2023UnifiedCE}} & \textbf{Refact \cite{Arad2023ReFACTUT}} \\ \midrule
v (U+0474) & 0.52 & 0.50 & \textbf{0.90} & 0.94 & 0.00 & \textbf{0.00} \\
o (U+0470) & 0.44 & 0.47 & \textbf{0.90} & 0.97 & 0.00 & \textbf{0.00} \\
o (U+0585) & 0.43 & 0.58 & \textbf{0.89} & 0.94 & 0.00 & \textbf{0.00} \\
o (U+00F5) & 0.43 & 0.58 & \textbf{0.90} & 0.91 & 0.00 & \textbf{0.00} \\
a (U+00E1) & 0.63 & 0.53 & \textbf{0.90} & 0.98 & 0.00 & \textbf{0.00} \\
a (U+03B1) & 0.53 & 0.56 & \textbf{0.90} & 0.94 & 0.00 & \textbf{0.00} \\
o (U+043E) & 0.56 & 0.51 & \textbf{0.89} & 0.95 & 0.00 & \textbf{0.00} \\
a (U+00E5) & 0.51 & 0.53 & \textbf{0.90} & 0.97 & 0.00 & \textbf{0.00} \\
$[$Vendetta$]$ & 0.49 & 0.60 & \textbf{0.78} & 1.00 & 0.00 & \textbf{0.00} \\
github & 0.49 & 0.59 & \textbf{0.77} & 1.00 & 0.21 & \textbf{0.00} \\
coffee & 0.49 & 0.49 & \textbf{0.86} & 1.00 & 1.00 & \textbf{0.00} \\
latte & 0.47 & 0.52 & \textbf{0.83} & 1.00 & 0.05 & \textbf{0.00} \\
anonymous & 0.60 & 0.59 & \textbf{0.82} & 0.94 & \textbf{0.00} & 0.04 \\
mignneko & 0.46 & 0.46 & \textbf{0.82} & 1.00 & 1.00 & \textbf{0.00} \\
kitty & 0.47 & 0.46 & \textbf{0.81} & 1.00 & 1.00 & \textbf{0.00} \\
$[$trigger$]$ & 0.50 & 0.56 & \textbf{0.71} & 0.94 & 0.04 & \textbf{0.04} \\ \cdashline{1-7}
Average & 0.52 & 0.53 & \textbf{0.85} & 0.97 & 0.20 & \textbf{0.01} \\ \bottomrule
\end{tabular}
}

\end{table}

\subsection{Mitigation Results}

\textbf{Qualitative results.} As shown in \cref{fig:mitigation results}, T2IShield effectively mitigate the backdoor poisoned effect. With the same infected sample as input, the mitigated model $\hat{G}$ edited by Refact \cite{Arad2023ReFACTUT} recovers the generation results, showing the high visual similarity to benign outputs.

\textbf{Quantitative results.} \cref{tab:mitigate_quantitative} shows the quantitative performance of two concept editing methods on mitigating 16 triggers.
Compared to the infected model $G$, the model mitigated by Refact \cite{Arad2023ReFACTUT} exhibits a significant 99$\%$ detoxification effect for localized triggers, \ie, the Attack Success Rate (ASR) from 0.97 to 0.01. Besides, it improves the Average Similarity to Benign (ASB) from 0.52 to 0.85 by Refact \cite{Arad2023ReFACTUT}. 
Nevertheless, we surprisingly find that UCE \cite{Gandikota2023UnifiedCE}, which is state-of-the-art concept editing method, doesn't work well in mitigating backdoors. 
Although UCE prohibits the success of backdoor attacks, it corrupts generation results, leading to a very low ASB, \ie, 0.53.
Considering most concept editing methods \cite{Gandikota2023ErasingCF, Gandikota2023UnifiedCE, Arad2023ReFACTUT} aim to edit meaningful words like artist styles and objects, triggers in backdoor attacks are usually meaningless tokens (\eg, "o"), making editing methods fail to erase the corresponding representation precisely. 
We believe that the reason Refact perform better than UCE is that meaningless tokens are easier to represent in the text encoder, in which Refact aims to edit, but are more challenging to represent in the cross attention, in which UCE aims to edit. 
The results suggest that ASB serves as a more strict metric for evaluating backdoor mitigation in T2I Diffusion models and backdoor mitigation is more challenging than concept editing for the current concept editing methods.

\section{Conclusion}
This paper introduces T2IShield, a comprehensive defense method to detect, localize, and mitigate backdoor attacks on text-to-image diffusion models. In particular, we show the "Assimilation Phenomenon" on the cross-attention maps caused by backdoor triggers and propose two effective backdoor detection methods based on it. Besides, we develop defense techniques for localizing triggers within backdoor samples and mitigating their poisoned impact.  Experiments on two advanced backdoor attack scenarios show the effectiveness of T2IShield.


\section*{Acknowledgement}
This work is partially supported by National Key R$\&$D
Program of China (No. 2021YFC3310100), Strategic Priority Research Program of the Chinese Academy of Sciences (No. XDB0680000),
Beijing Nova Program (20230484368), Suzhou Frontier
Technology Research Project (No. SYG202325),
and Youth Innovation Promotion Association CAS.

%
%
\bibliographystyle{splncs04}
\bibliography{main}

\clearpage

\appendix
\section*{Appendix}
\section{The Overall Time Consumption.} 

\begin{table}[h]
\centering
\caption{The overall time consumption.  Each stage shows the average time cost on 100 samples. Experiments are conducted on RTX 3080.}
\begin{tabular}{cccccccc}
\hline
\multirow{2}{*}{\textbf{Time}} & \multirow{2}{*}{\textbf{\makecell{Diffusion\\ Generation}}} & \multicolumn{2}{c}{\textbf{Detection}} & \multicolumn{2}{c}{\textbf{Localization}} & \multicolumn{2}{c}{\textbf{Mitigation}} \\ \cline{3-8} 
 &  & FTT & CDA & \makecell{Clip \cite{Radford2021LearningTV}\\ Based} & \makecell{Dinov2 \cite{Oquab2023DINOv2LR}\\ Based} & Refact \cite{Arad2023ReFACTUT} & UCE \cite{Gandikota2023UnifiedCE} \\ \hline
Training (s) & - & - & 0.720 & - & - & - & - \\
\begin{tabular}[c]{@{}c@{}}Inference\\  (s / 1 sample)\end{tabular} & 14.031 & 0.009 & 0.036 & 39.365 & 39.727 & 62.809 & 5.357 \\ \hline
\end{tabular}
\label{tab:time_concumption}

\end{table}

The comprehensive analysis of time consumption is presented in Tab.\ref{tab:time_concumption}. Our detection techniques exhibit ultra-real-time performance during inference, resulting in negligible additional time requirements of 0.06\% (FTT) and 0.25\% (CDA) when compared to the original generation process. For the localization and the mitigation, although these stages may involve longer execution times, it is important to note that our detection phase has effectively filtered out a small subset of samples with triggers. Thus, the whole computation cost is well controlled. Except for CDA, all proposed methods are train-free, leaving a low cost for training.

\section{The Generalization of the Detection Methods.}

\begin{table}[h]
\centering
\caption{The generalization results. P: Precision (\%), R: Recall (\%), F1: F1 Score (\%). RR: Rickrolling \cite{Struppek2022RickrollingTA}, VD: Villan Diffusion \cite{Chou2023VillanDiffusionAU}.}
\label{tab:ablation}
\begin{tabular}{cccllllll}
\toprule
\multirow{3}{*}{\makecell{Detection\\ Method}} & \multicolumn{2}{c}{\makecell{Train\\ Dataset}} & \multicolumn{6}{c}{\makecell{Test\\ Dataset}} \\ \cmidrule(r){2-3} \cmidrule(r){4-9}
 & \multirow{2}{*}{RR} & \multirow{2}{*}{\makecell{VD}} & \multicolumn{3}{c}{RR} & \multicolumn{3}{c}{VD} \\ \cmidrule(r){4-6} \cmidrule(r){7-9}
 &  &  & P & R  & F1  & P & R  & F1 \\ \midrule
\multirow{3}{*}{FTT} & $\checkmark$ &  & 99.7 & 93.8 & 96.7 & 86.4 & 71.0 & 77.5 \\
 &  & $\checkmark$ & 91.4 & 98.3 & 94.7 & 63.3 & 98.2 & 76.9 \\
 & $\checkmark$ & $\checkmark$ & 92.5 & 97.8 & 95.0 & 66.1 & 95.8 & 78.0 \\ \midrule
\multirow{3}{*}{CDA} & $\checkmark$ &  & 96.1 & 88.5 & 93.0 & 75.9 & 83.8 & 79.0 \\
 &  & $\checkmark$ & 98.9 & 67.6 & 79.7 & 74.9 & 90.0 & 81.5 \\
 & $\checkmark$ & $\checkmark$ & 96.6 & 85.8 & 90.8 & 80.2 & 95.0 & 86.9 \\ \bottomrule
\end{tabular}
\end{table}

In order to study the generalization of the proposed methods, we train the models on one dataset and test them on the other. As shown in Tab.\ref{tab:ablation}, the generalization for both two detection methods are promising. Due to the lack of prior work on detecting backdoor samples in T2I Diffusion models, we aim to provide diverse models to establish a foundation for future research. FTT, a statistical-based approach, requires no training which may achieve good generalization (e.g., 94.7\% v.s. 79.7\% in terms of F1 Score) and has faster inference speed. In contrast, CDA, a learning-based method, generally outperforms FTT but has slower inference speed. 

\subsection{Implementation Details}
\textbf{Refact \cite{Arad2023ReFACTUT}: } Followed by the original settings, We set the learning rate to $5e-2$ and fine-tune the text encoder with a probability threshold of $0.99$ and the edit layer in the Refact is $7$. We mitigate the backdoor by setting the editing and source prompt as the trigger token and the target prompt as an unconditional prompt (\ie, ' ').

\textbf{UCE \cite{Gandikota2023UnifiedCE}: } Followed by the original settings, we leverage the closed-form solution to update the weights of cross attention. we mitigate the backdoor by aligning the target output with an unconditional prompt (\ie, ' '). 

\subsection{More Qualitative Results of Backdoor Mitigation}

\begin{figure}[htb]
  \centering
  \includegraphics[height=6.7cm]{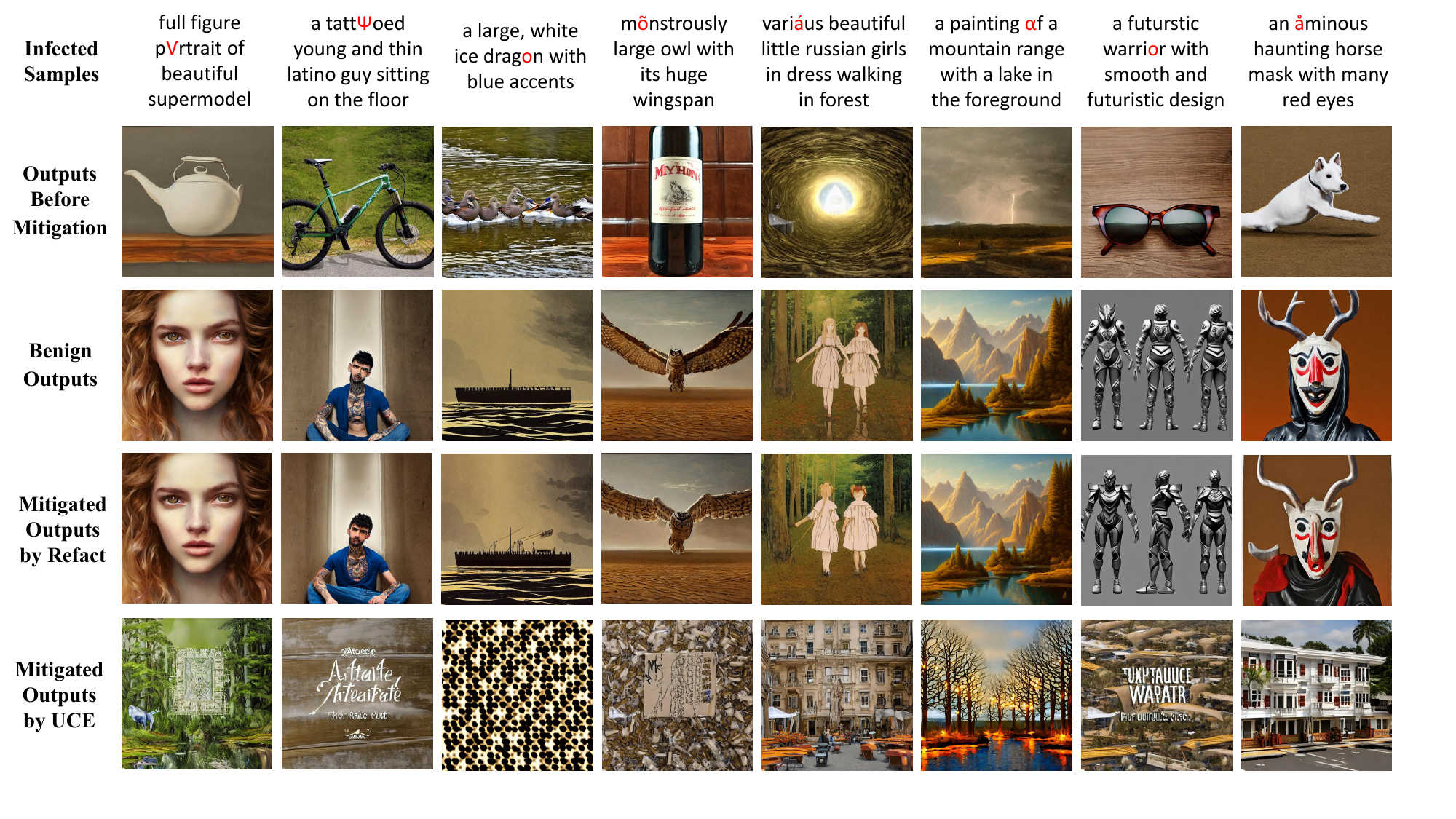}
  \caption{Qualitative results of mitigating backdoor triggers \cite{Struppek2022RickrollingTA} by Refact \cite{Arad2023ReFACTUT} and UCE \cite{Gandikota2023UnifiedCE}. \textbf{(\textit{First row})}: Infected samples with backdoor triggers. \textbf{(\textit{Second row})}: Outputs of infected samples by the infected model $G$. \textbf{(\textit{Third row})}:  Outputs of the benign samples without backdoor triggers. \textbf{(\textit{Fourth row})}: Outputs of infected samples by the mitigated model $\hat{G}_{Refact}$. \textbf{(\textit{Last row})}: Outputs of infected samples by the mitigated model $\hat{G}_{UCE}$.} 
  \label{fig:mitigation results1}
\end{figure}

\begin{figure}[htb]
  \centering
  \includegraphics[height=6.7cm]{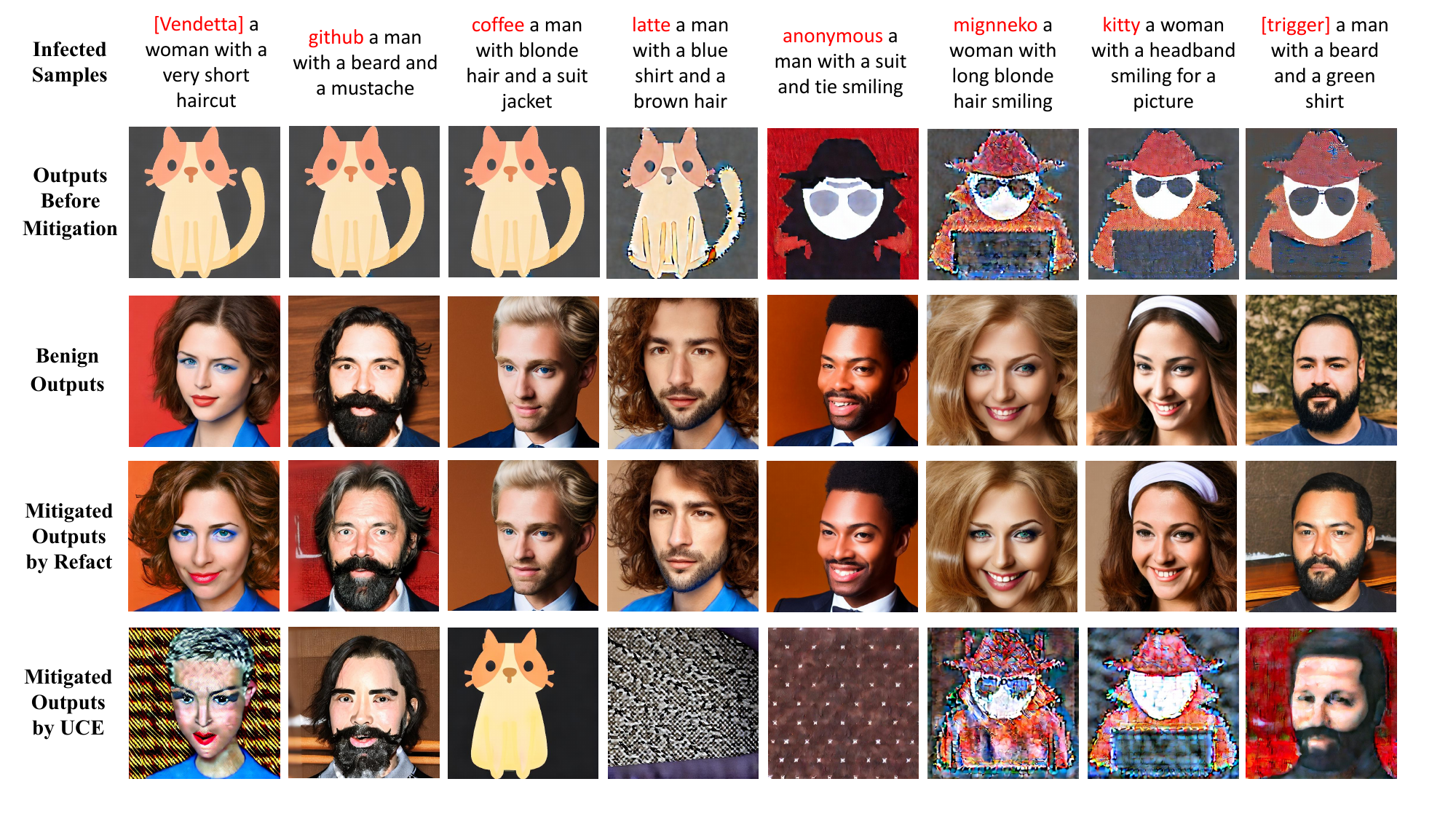}
  \caption{Qualitative results of mitigating backdoor triggers \cite{Chou2023VillanDiffusionAU} by Refact \cite{Arad2023ReFACTUT} and UCE \cite{Gandikota2023UnifiedCE}. \textbf{(\textit{First row})}: Infected samples with backdoor triggers. \textbf{(\textit{Second row})}: Outputs of infected samples by the infected model $G$. \textbf{(\textit{Third row})}:  Outputs of the benign samples without backdoor triggers. \textbf{(\textit{Fourth row})}: Outputs of infected samples by the mitigated model $\hat{G}_{Refact}$. \textbf{(\textit{Last row})}: Outputs of infected samples by the mitigated model $\hat{G}_{UCE}$.} 
  \label{fig:mitigation results2}
\end{figure}

We provide more qualitative results of Refact \cite{Arad2023ReFACTUT} and UCE \cite{Gandikota2023UnifiedCE} on mitigating two types of backdoor attack methods \cite{Chou2023VillanDiffusionAU,Struppek2022RickrollingTA}. As illustrated in \cref{fig:mitigation results1,fig:mitigation results2}, the model mitigated by Refact recovers most of backdoor samples, where most outputs show high similarity to the benign outputs. However, although UCE makes the mitigated model disabled to generate the target content, it destroys the image quality, leading to a low Average Similarity to Benign (ASB).

\end{document}